\documentclass[letterpaper]{article} % DO NOT CHANGE THIS
\usepackage{aaai24}  % DO NOT CHANGE THIS
\usepackage{times}  % DO NOT CHANGE THIS
\usepackage{helvet}  % DO NOT CHANGE THIS
\usepackage{courier}  % DO NOT CHANGE THIS
\usepackage[hyphens]{url}  % DO NOT CHANGE THIS
\usepackage{graphicx} % DO NOT CHANGE THIS
\usepackage{natbib}  % DO NOT CHANGE THIS AND DO NOT ADD ANY OPTIONS TO IT
\usepackage{caption} % DO NOT CHANGE THIS AND DO NOT ADD ANY OPTIONS TO IT
\usepackage[ruled, noend, noline]{algorithm2e}

\usepackage{newfloat}
\usepackage{listings}
\DeclareCaptionStyle{ruled}{labelfont=normalfont,labelsep=colon,strut=off} % DO NOT CHANGE THIS
\usepackage{amsmath}
\DeclareMathAlphabet{\mathbbold}{U}{bbold}{m}{n}
\newcommand*{\boldone}{\mathbbold{1}}
\usepackage{accents}
\newcommand{\ubar}[1]{\underaccent{\bar}{#1}}
\usepackage{booktabs}

\usepackage{color}

\title{Optimal Survival Trees\\A Dynamic Programming Approach}
\author{
    Tim Huisman, 
    Jacobus G. M. van der Linden, 
    Emir Demirovi\'{c}
}
\affiliations{
    Delft University of Technology\\
    T.J.Huisman-1@student.tudelft.nl, \{J.G.M.vanderLinden, E.Demirovic\}@tudelft.nl
}

\DeclareRobustCommand{\VAN}[3]{#2} % set up for citation

\begin{document}

\maketitle

\begin{abstract}
Survival analysis studies and predicts the time of death, or other singular unrepeated events, based on historical data, while the true time of death for some instances is unknown. Survival trees enable the discovery of complex nonlinear relations in a compact human comprehensible model, by recursively splitting the population and predicting a distinct survival distribution in each leaf node.
We use dynamic programming to provide the first survival tree method with optimality guarantees, enabling the assessment of the optimality gap of heuristics. We improve the scalability of our method through a special algorithm for computing trees up to depth two.
The experiments show that our method's run time even outperforms some heuristics for realistic cases while obtaining similar out-of-sample performance with the state-of-the-art.
\end{abstract}

\section{Introduction}
% Intro Survival Analysis
The aim of \emph{survival analysis} is to study and predict the time until a singular unrepeated event occurs, such as death or mechanical failure.
% Broad applicability of survival analysis
Applications include not only evaluating the effectiveness of medical treatment \citep{selvin2008sa_medical_book}, but also, for example, recidivism risk estimation in criminology \citep{chung1991sa_survey_criminology}, fish migration analysis \citep{castro_santos2003sa_fish_migration} and studies on human migration and fertility \citep{eryurt2012sa_turkey_migration_fertility}.

% Censored data
Survival analysis is challenging because the time of event of some instances is unknown, i.e., it is \emph{censored}. This study focuses on the most common form of censoring: \emph{right-censored} data, where the true time of the event is unknown, for example, because an instance survived beyond the end of the experiment record.

\begin{figure}[b!]
    \centering
    \includegraphics[width=0.95\columnwidth]{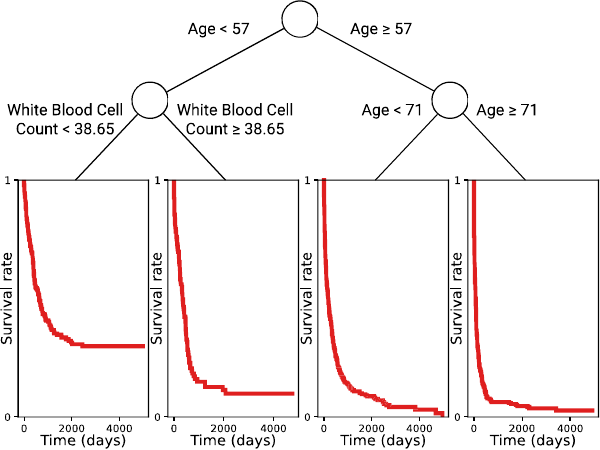} 
    \caption{An example of a survival tree. Each leaf has a different survival distribution.}
    \label{fig:survival_tree_example}
\end{figure}

% Decision trees for survival analysis, motivated by interpretability
The application of survival analysis in the medical and other high-stake domains motivates the use of human-interpretable machine learning models, such as \emph{decision trees} \citep{rudin2019stop_black_box}.
A decision tree recursively partitions instances by their attributes into buckets, i.e., the leaves of the tree. In each of these leaf nodes, a survival curve can be computed, as shown, for example, in Fig.~\ref{fig:survival_tree_example}.
The advantage of decision trees is that they can model complex nonparametric relations while remaining easy to understand \citep{freitas2014comprehensible, carrizosa2021mathematical}. 
\citet{davis1989exponential_survival} provided one of the first survival tree methods, by applying recursive splitting of censored survival data with an interface similar to CART \citep{breiman1984cart}. 
Other similar greedy top-down induction heuristics were developed by \citet{leblanc1993surv_tree_goodness_split}, \citet{su2004multivar_surv_trees_maxlikeli},
and \citet{hothorn2006survival_distance}, each applying different splitting techniques.

% Near-optimal Survival trees OST
To improve the performance of survival trees, \citet{bertsimas2022survival} recently proposed a local search method called Optimal Survival Trees (OST), based on the method proposed in \citep{dunn2018thesis, bertsimas2019book_ml_opt_lens}. Despite its name, OST does not provide a guarantee of global optimality, but iteratively finds local improvements to the tree structure and converges to a local optimum. 

% Optimal decision trees motivation
Trees that do provide a guarantee of global optimality over a training set for a given size limit are called optimal decision trees.
Out-of-sample results for optimal decision trees for classification typically also show an improvement over heuristics \citep{bertsimas2017optimal}. Therefore, optimal decision tree methods can provide better performance, while producing smaller trees, which increases their interpretability \citep{piltaver2016comprehensible}. However, to the best of our knowledge, there are no globally optimal decision tree methods for survival analysis yet.

% Scalability challenge
The challenge of finding optimal decision trees is scalability since it is an NP-Hard problem \citep{laurent1976constructing}. Therefore, many optimal decision tree methods that optimize accuracy lack scalability. This includes methods based on mixed-integer programming \citep{bertsimas2017optimal}, constraint programming \citep{verhaeghe2020cp_odt} and Satisfiability \citep{narodytska2018sat_odt}.

% Promising results in optimal decision trees using dynamic programming
Better scalability results have been obtained by using dynamic programming (DP) because it directly exploits the recursive structure of the trees 
by treating subtrees as independent subproblems \citep{demirovic2022murtree}. \citet{linden2023streed} show that these results also generalize beyond maximizing accuracy.

% Contribution
Our contributions are a first optimal survival tree algorithm called \emph{SurTree}, based on a dynamic programming approach; an algorithm for trees of maximum depth two that greatly improves scalability; and a detailed experimental comparison of our new method with the local search method OST and the greedy heuristic CTree \citep{hothorn2006survival_distance}. The first two contributions are inspired by the scalability improvements obtained for optimal classification trees with DP by \citet{demirovic2022murtree}.
Our experiments show that SurTree's out-of-sample performance on average is better than CTree and similar to OST while outperforming OST in run time for realistic cases. Since SurTree is the first optimal method for survival trees, our method also helps assess the quality of heuristic solutions.

% Paper setup
The following sections introduce related work and the preliminaries for our work. We then present our DP-based approach to optimal survival trees, evaluate it on synthetic and real data sets, and compare it with the state-of-the-art.

\section{Related Work}
%This section gives further background and related work on survival analysis, survival trees, and optimal decision trees.

\paragraph{Survival analysis}
Traditionally, many statistical approaches have been developed for dealing with censored data \citep{chung1991sa_survey_criminology}. This includes nonparametric approaches, such as the Kaplan-Meier method \citep{kaplan1958kaplan_meier} and the Nelson-Aalen estimator \citep{nelson1972nelson_aalen_estimator, aalen1978nelsen_aalen_estimator}, semiparametric approaches, such as Cox proportional hazards regression \citep{cox1972sa_reg}, and parametric approaches such as linear regression.
\citet{wang2019sa_review} provide an overview of the later use of machine learning for survival analysis, including survival trees, random survival forests \citep{ishwaran2008random_survival_forests}, support vector machines \citep{vanbelle2011sa_svm}, and neural networks \citep{chi2007sa_nn}.

\paragraph{Survival trees}
Since computing optimal decision trees is NP-Hard \citep{laurent1976constructing}, traditionally most decision tree methods use \emph{greedy top-down induction} based on some \emph{splitting criterion}, such as Gini impurity or information gain \citep{breiman1984cart, quinlan1993c4.5}. 
For survival trees, many such splitting criteria have been proposed. They can be divided into criteria that promote within-node homogeneity or between-node heterogeneity. Examples of the former are \citep{gordon1985survival_node_purity, davis1989exponential_survival, therneau1990survival_martingale} and \citep{leblanc1992sa_rel_risk}. Examples of the latter are \citep{ciampi1986survival_distance,segal1988sa_reg_trees} and \citep{leblanc1993surv_tree_goodness_split}. 
Other developments are presented by \citet{molinaro2004survival_ipcw}, who propose to weigh the uncensored data based on inverse-propensity weighting; \citet{su2004multivar_surv_trees_maxlikeli}, who consider survival analysis for clustered events using a maximum likelihood approach based on frailty models; and \citet{hothorn2006survival_distance}, who introduce $\chi^2$ tests as stopping criterion to prevent variable selection bias.

Recently, \citet{bertsimas2022survival} presented OST (optimal survival trees), based on the coordinate-descent method proposed by \citet{dunn2018thesis}, by iteratively improving one node in the tree until a local optimum is reached. Because the problem is non-convex, they repeat this process several times to increase the probability of finding a good tree. Their results show that local search can outperform greedy heuristics such as \citep{therneau1990survival_martingale} and \citep{hothorn2006survival_distance}. However, despite its name, OST provides no guarantee of converging to the global optimum.

\paragraph{Optimal decision trees}
A popular approach for computing optimal decision trees is the use of general-purpose solvers. \citet{bertsimas2017optimal} and \citet{verwer2017flexible} showed how optimizing decision trees can be formulated as a mixed-integer programming (MIP) formulation. These MIP formulations were later improved by several others \citep{verwer2019learning, zhu2020scalable, aghaei2021strong}.
\citet{verhaeghe2020cp_odt} showed how constraint programming can be used to optimize trees. \citet{narodytska2018sat_odt} and \citet{janota2020dt_sat_expl_paths} presented a Satisfiability (SAT) approach for finding a perfect minimal tree. \citet{hu2020maxsat_odt} developed a maximum Satisfiability (MaxSAT) approach, while \citet{shati2021sat_dt_nonbin} extended the use of MaxSAT beyond binary predicates. However, as of yet, each of these approaches struggles to scale beyond small data sets and tree-size limits.

Recently, major improvements in scalability have been obtained using a dynamic programming (DP) approach, which as a result often outperforms the MIP, CP and (Max)SAT approaches by several orders of magnitude \citep{aglin2020learning, demirovic2022murtree, linden2023streed}. \citet{nijssen2007mining} were one of the first to propose the use of DP for optimizing decision trees. \citet{nijssen2010dl8_constraints} showed how DP can also be used to optimize other objectives than accuracy, provided the objective is additive. \citet{hu2019sparse, lin2020generalized_sparse, aglin2020learning, aglin2020pydl8} and \citet{demirovic2022murtree} improved the DP approach by introducing branching and bounding, new forms of caching, better bounds, and special procedures for trees of depth two. \citet{linden2023streed} prove that this DP approach can be applied to any separable optimization task, i.e., an optimization problem that can be independently solved for subtrees.

Since we do not know of any optimal survival tree method, and given the success of DP methods mentioned above, this study explores the use of DP for survival trees and the benefit of globally optimizing survival trees.

\section{Preliminaries}
%This section introduces all notation, terminology, and background information, and provides a formal description of survival analysis.

\paragraph{Definitions and notation}
\label{section:prelim_terminology}
We aim to optimize a survival tree over a data set $\mathcal{D}$. This data set consists of instances that either experienced the event of interest or were censored. Each of these instances is described by a set of features. The following defines each of these terms:

An \emph{event of interest}, or simply \emph{event} or \emph{death}, is the non-repeatable event for which the time until occurrence is measured within a trial. The \emph{time-to-event} is the amount of time leading up to the event of interest from the beginning of the observation.
In this study, we consider \emph{right-censored} data, which means that for some instances the exact time-to-event is unknown, but lower bounded by some known time, for example, because a patient left the trial before the event of interest could be observed.

A data set $\mathcal{D}$, or a trial, consists of a set of \emph{instances} $(t_i, \delta_i, \textbf{fv}_i)$, each described by a feature vector $\textbf{fv}_i$, a \emph{censoring indicator} $\delta_i \in \{0, 1\}$ stating whether the event of interest was observed and a \emph{time} $t_i > 0$.
In case of censoring ($\delta_i = 0$), $t_i$ denotes the last time of observation.
Otherwise, $t_i$ denotes the time-to-event.
The feature vector describes the instance by a set of features $\mathcal{F}$. Our method assumes that all features are binarized beforehand such that each feature is a binary predicate. We write $f \in \textbf{fv}_i$ if the predicate holds for instance $i$ or $\bar{f} \in \textbf{fv}_i$ if it does not hold.
We use $\mathcal{D}(f_i)$ and $\mathcal{D}(\bar{f_i})$ to refer to all instances in $\mathcal{D}$ for which the predicate $f_i$ is valid or not, respectively. Multiple feature splits can be stacked so that, for example, $\mathcal{D}(f_1, \bar{f_2})$ refers to all instances for which $f_1$ holds and $f_2$ does not hold.

A \emph{decision tree} partitions instances based on their features. We consider \emph{binary} trees, where each node is either a \emph{decision node} with two children, or a \emph{leaf node}. Each decision node splits the data set on a certain feature. Each leaf node assigns a label to every instance that ends up at that leaf node.
A \emph{survival tree} (see Fig. \ref{fig:survival_tree_example}) is a special type of decision tree that assigns in each leaf node not just a label, but a survival distribution that describes the survival odds after a certain amount of time.

\paragraph{Survival analysis background}
The goal of survival analysis is to accurately describe the \emph{survival function}, which gives the probability of survival after a time $t$, denoted as $S(t) = P(T \geq t)$, with $T$ the true time of the event \citep{wang2019sa_review}. Its opposite is the cumulative death distribution function $F(t) = 1 - S(t)$, with its derivative, the death density function $f(t) = \frac{d}{dt} F(t)$.

One of the most used estimators for the survival function is the Kaplan-Meier estimator \citep{kaplan1958kaplan_meier}:
\begin{equation}
    \hat{S}(t) = \prod_{t' \leq t} \bigg(1 - \frac{d(t')}{n(t')} \bigg)
\end{equation}%
with $d(t')$ the number of deaths at time $t'$ and $n(t')$ the number of survivors up and until time $t'$:
\begin{align}
d(t) &= |\{ (t_i, \delta_i, \textbf{fv}_i) \in \mathcal{D} ~|~ t_i = t \land \delta_i = 1 \} | \\
n(t) &= |\{ (t_i, \delta_i, \textbf{fv}_i) \in \mathcal{D} ~|~ t_i \geq t \} |
\end{align}

The \emph{hazard function} (also known by the force of mortality, the instantaneous death rate, or the conditional failure rate), given by $\lambda(t) = \frac{f(t)}{S(t)}$, indicates the frequency (or rate) of the event of interest happening at time $t$, provided that it has not happened before time $t$ yet \citep{dunn2009basic_statistics}. Alternatively, it can be written as $\lambda(t) = - \frac{d}{dt} \ln{S(t)}$. The \emph{cumulative hazard function} is the integral over the hazard function $\Lambda(t) = \int_0^t \lambda(u)du$, and thus the survival function $S(t)$ can be rewritten as:
\begin{equation}
\label{eq:surv_func_exp_cum_hazard}
    S(t) = e^{-\Lambda(t)}
\end{equation}

A commonly used estimator for the cumulative hazard function is the Nelson-Aalen estimator, which is defined analogously to the Kaplan-Meier estimator \citep{nelson1972nelson_aalen_estimator, aalen1978nelsen_aalen_estimator}:
\begin{equation}
\label{eq:nelson_aalen_estimator}
    \hat{\Lambda}(t) = \sum_{t' \leq t} \frac{d(t')}{n(t')}
\end{equation}
The Nelson-Aalen estimator of Eq.~\eqref{eq:nelson_aalen_estimator} in combination with Eq.~\eqref{eq:surv_func_exp_cum_hazard} is what we will use for our method, as explained in the next section.

\section{Method}
We present \emph{SurTree}, a dynamic programming approach to optimizing survival trees. First, we explain what loss function is minimized. Second, we show how DP can be used to find the global optimum for the loss function. Third, we present a special algorithm for trees of depth two that results in a significant increase in scalability.

\subsection{Loss function}
\begin{figure}[t]
    \centering
    \includegraphics[width=\columnwidth]{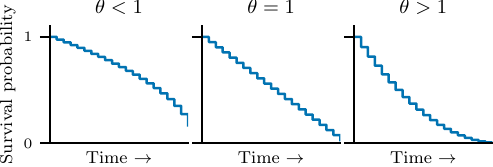}
    \caption{A visualization of how $\theta$ affects a survival distribution $\hat{S}(t)$. Every plot uses the same $\hat\Lambda(t)$, but use $\theta = 0.5$, $\theta = 1$ and $\theta = 2$ respectively.}
    \label{fig:theta_example}
\end{figure}
The optimization of decision trees requires a target loss function.
For computational efficiency, the loss function over a leaf node needs to be independent of other leaf nodes.
Therefore, like \citep{bertsimas2022survival}, we optimize the likelihood method from \citep{leblanc1992sa_rel_risk}.
This method assumes that the survival function $S_i$ for each instance $i$ can be approximated by a \emph{proportional hazard model}, described by multiplying the exponent in Eq.~\eqref{eq:surv_func_exp_cum_hazard}, that is, the baseline hazard model $\hat{\Lambda}(t)$, as estimated by the Nelson-Aalen estimator in Eq.~\eqref{eq:nelson_aalen_estimator}, by some parameter $\theta_i$:
\begin{equation}
    \hat{S}_i(t) = e^{-\theta_i\hat{\Lambda}(t)}
\end{equation}
Fig.~\ref{fig:theta_example} shows how $\theta_i$ changes the survival function $S_i(t)$.

\citet{leblanc1992sa_rel_risk} show that for a given data set $\mathcal{D}$ the estimate $\hat{\theta}$ with maximum likelihood is equal to:
\begin{equation}
    \label{eq:surtree_theta_function}
    \hat{\theta} = \frac{\sum_{(t_i, \delta_i, \textbf{fv}_i) \in \mathcal{D}} \delta_i}{\sum_{(t_i, \delta_i, \textbf{fv}_i) \in \mathcal{D}} \hat{\Lambda}(t_i)}
\end{equation}
This means that for a single instance $i$, the saturated coefficient that perfectly maximizes the likelihood for that instance alone is given by:
\begin{equation}
\hat{\theta}^{sat}_i = \frac{\delta_i}{\hat{\Lambda}(t_i)}
\end{equation}%
The loss for a single instance is then defined as the difference between the log-likelihood of the instance's leaf node $\hat{\theta}$ and the log-likelihood of the instance's $\hat{\theta}^{sat}_i$. In the appendix, we show how this results in the following loss for a data set $\mathcal{D}$ that ends up in a leaf node with parameter $\hat{\theta}$:
\begin{multline}
\label{eq:surtree_error_function}
\mathcal{L}(\mathcal{D}, \hat{\theta}) = \\
\sum_{(t_i, \delta_i, \textbf{fv}_i) \in \mathcal{D}} \Big( \hat\Lambda(t_i) \hat{\theta} -\delta_i\log\hat\Lambda(t_i) - \delta_i\log\hat{\theta} - \delta_i \Big)
\end{multline}

\subsection{Dynamic Programming Approach}
The loss function of Eq.~\eqref{eq:surtree_error_function} consists of several nonlinear terms that prevent it from being directly optimized with mixed-integer linear programming. However, it can be optimized with dynamic programming. The key change compared to a DP formulation for standard decision trees is the base case: instead of assigning a class based on the majority vote, we now optimize $\hat{\theta}$ such that the loss is minimized. We apply this to the DP formulation from \citep{demirovic2022murtree}:
\begin{multline}
\label{eq:surtree_recurrence}
T(\mathcal{D}, d, n) = \\
    \begin{cases} 
        \min_{\hat{\theta}} \mathcal{L}(\mathcal{D}, \hat{\theta}) & n = 0  \\
        T(\mathcal{D}, d, 2^d - 1) & n > 2^d - 1 \\
        T(\mathcal{D}, n, n) & d > n \\
        \min\{T(\mathcal{D}(\overline{f}), d - 1, n - i - 1) \\ ~~~ + T(\mathcal{D}(f), d - 1, i) \\ ~~~ : f \in \mathcal{F}, i \in [0, n - 1]\} & \text{otherwise} \\
    \end{cases}
\end{multline}

In this equation, subproblems are defined by the dataset $\mathcal{D}$, the (remaining) tree depth $d$, and branching node budget $n$. When $n=0$, a leaf node is returned with a survival distribution given by $\theta$ for which the loss is minimized. When the depth or branching node budget exceeds what is possible according to the other budget, for example, when $d > n$, the budgets are updated accordingly. Otherwise, a branching node is optimized by looping over all possible branching features $f\in\mathcal{F}$ and all possible branching node budget distributions. The loss of the two subtrees is summed for each possible split, and the best possible split is returned.

The solutions to the subproblems $\langle \mathcal{D}, d, n \rangle$ are \emph{cached}. Cached solutions are also used as lower bounds. Upper bounds (best solution so far) and lower bounds for a subtree search are used to terminate the search early.

To prevent overfitting, we use \emph{hyper-tuning} to tune the depth and number of branching nodes. Alternatively, a cost-complexity parameter can be used to penalize adding more branching nodes. However, tuning the depth and number of nodes directly allows to reuse the cache, yielding a speed improvement, without loss of solution quality.

\subsection{Trees of Depth Two}
\citet{demirovic2022murtree} developed a major scalability improvement for optimizing classification trees of maximum depth two. Instead of applying the splitting and recursing technique (as done similarly in Eq.~\eqref{eq:surtree_recurrence}), which requires counting class occurrences for every possible leaf node, this algorithm precomputes the class occurrences by looping over all pairs of features in the feature vector $f_i, f_j \in \textbf{fv}_k$ for each instance $k$. The counts can then be used to directly compute the misclassification score for each leaf node without having to examine the entire data set again. 
\citet{linden2023streed} show that this method can also be generalized to other optimization tasks, provided that a per-instance breakdown of the loss can be formulated.

Here, we provide a breakdown of the per-instance contribution to the costs, such that the same precomputing technique can be used for survival analysis. Pseudocode is provided in the appendix.

First, we split Eq.~\eqref{eq:surtree_error_function} into several summations:
\begin{equation}
    \hat{\theta} \sum_i \hat\Lambda(t_i)
    - \sum_i \delta_i\log\hat\Lambda(t_i) 
    - \log\hat{\theta} \sum_i \delta_i 
    - \sum_i \delta_i    
\end{equation} 

Then, by substituting Eq.~\eqref{eq:surtree_theta_function} into the above formula, we get the following:
%\begin{equation}
\begin{align}
\mathcal{L}(\mathcal{D}, \hat{\theta}) =& 
\frac{\sum_i \delta_i}{\sum_i \hat\Lambda(t_i)} \sum_i \hat\Lambda(t_i)
- \sum_i \delta_i\log\hat\Lambda(t_i) \nonumber \\
\label{eq:redefinition_error_function}
& - \log\Bigg(\frac{\sum_i \delta_i}{\sum_i \hat\Lambda(t_i)}\Bigg) \sum_i \delta_i
- \sum_i \delta_i  \\
=& - \sum_i \delta_i\log\hat\Lambda(t_i) - \log\Bigg(\frac{\sum_i \delta_i}{\sum_i \hat\Lambda(t_i)}\Bigg) \sum_i \delta_i \nonumber
    \end{align}
%\end{equation}

Eq.~\eqref{eq:redefinition_error_function} is expressed as a function of several sums over the instances. These sums can be precomputed in the same way as class occurrences are precomputed in \citep{demirovic2022murtree}.
Three sums need to be computed: the \textit{event sum} $\operatorname{ES}$, the \textit{hazard sum} $\operatorname{HS}$, and the \textit{negative log hazard sum} $\operatorname{NLHS}$.
\begin{align}
    \label{eq:surtree_es}
    \operatorname{ES}(f_i, f_j) &= \sum_{(t_k, \delta_k, \textbf{fv}_k) \in \mathcal{D}(f_i, f_j)} \delta_k \\
    \label{eq:surtree_hs}
    \operatorname{HS}(f_i, f_j) &= \sum_{(t_k, \delta_k, \textbf{fv}_k) \in \mathcal{D}(f_i, f_j)} \hat\Lambda(t_k) \\
    \label{eq:surtree_nlhs}
    \operatorname{NLHS}(f_i, f_j)  &= \sum_{(t_k, \delta_k, \textbf{fv}_k) \in \mathcal{D}(f_i, f_j)} -\delta_k\log\hat\Lambda(t_k)
\end{align}

Eqs.~\eqref{eq:surtree_es}-\eqref{eq:surtree_nlhs} compute the event, hazard, and negative log hazard sum for the leaf node with data that satisfies feature $f_i$ and $f_j$. The sums for the other leaf nodes can be computed as follows (similarly for  $\operatorname{HS}$ and $\operatorname{NLHS}$):
\begin{align}
    \label{eq:murtree_es_ni}
    \operatorname{ES}(\overline{f_i}) &= \operatorname{ES} - \operatorname{ES}(f_i) \\
    \label{eq:murtree_es_i_nj}
    \operatorname{ES}(f_i, \overline{f_j}) &= \operatorname{ES}(f_i) - \operatorname{ES}(f_i, f_j) \\
    \label{eq:murtree_es_ni_j}
    \operatorname{ES}(\overline{f_i}, f_j) &= \operatorname{ES}(f_j) - \operatorname{ES}(f_i, f_j) \\
    \label{eq:murtree_es_ni_nj}
    \operatorname{ES}(\overline{f_i}, \overline{f_j}) &= \operatorname{ES} - \operatorname{ES}(f_i) - \operatorname{ES}(f_j) + \operatorname{ES}(f_i, f_j)
\end{align}

Here, $\operatorname{ES}$ denotes the event sum over the whole dataset $\mathcal{D}$, while $\operatorname{ES}(f_i)$ denotes the event sum on the dataset $\mathcal{D}(f_i)$. Once these sums are computed for pairs of features, the final loss for each split and each possible leaf node can be computed from the sums:
\begin{equation}
    \label{eq:surtree_terminal_error}
    \mathcal{L}(\mathcal{D}) = \operatorname{NLHS} - \operatorname{ES} \log\bigg(\frac{\operatorname{ES}}{\operatorname{HS}}\bigg)
\end{equation}

Since we only explicitly count the values for when $f_i$ and $f_j$ hold, and derive the other cases implicitly through Eqs.~\eqref{eq:murtree_es_ni}-\eqref{eq:murtree_es_ni_nj}, the run time is reduced from $O(|\mathcal{F}|^2 |\mathcal{D}|)$ to $O(m^2 |\mathcal{D}|)$, with $m$ the maximum number of features that hold for any instance in $\mathcal{D}$. This is specifically advantageous when features hold sparingly. Non-sparse features are flipped to improve sparsity. 

\section{Experiments}
The following introduces the experiment setup, the survival analysis metrics, a scalability analysis with an evaluation of the impact of our depth-two algorithm, and the out-of-sample performance of SurTree and two other methods.
%First, we provide the experiment setup. Second, we introduce the metrics that are used for evaluation. Third, we give a scalability analysis of SurTree and the impact of our depth-two algorithm. Finally, we compare the out-of-sample performance of all methods for both synthetic and real data. 

\subsection{Experiment Setup}
\paragraph{Methods} We implemented SurTree in C++ with a Python interface using the STreeD framework \citep{linden2023streed}.\footnote{\url{https://github.com/AlgTUDelft/pystreed}} 
In our experiment setup,\footnote{\url{https://github.com/TimHuisman1703/streed_sa_pipeline}} we compare SurTree with the Julia implementation of OST \citep{bertsimas2022survival} and the R implementation of CTree \citep{hothorn2006survival_distance}. 
Each method is tuned using ten-fold cross-validation. For SurTree, we tune the depth and node budget. For CTree, we tune the confidence criterion. For OST, we tune the depth and, simultaneously, OST automatically tunes the cost-complexity parameter as part of its training.
All experiments were run on an Intel i7-6600U CPU with 4GB RAM with a time-out of 10 minutes. 

\paragraph{Data} We evaluate both on synthetic data to measure the effect of censoring and of having more data, and on real data sets. The real data sets are taken from the SurvSet repository \citep{drysdale2022survset}. Since the results on the synthetic data show that the differences between the methods are more clearly visible for larger datasets, we limit our real data analysis to data sets with more than 2000 instances. We evaluate out-of-sample performance on the real data sets using five-fold cross-validation. 

The synthetic data is generated according to the procedure described in \citep{bertsimas2022survival}.
First, we generate $n$ feature vectors, with three continuous features, one binary feature, and two categorical features with three and five categories. Each of the features is uniformly distributed.
Second, we randomly generate a survival tree $T$ of depth five that splits on random features and assign a random distribution to each leaf node (see the appendix for a list of used distributions).
Third, for each of the $n$ instances, we classify the instance using the tree and assign it a random time-to-event $t_i$ by sampling from the corresponding leaf distribution. After that, we assign the instance a random value $u_i$, uniformly distributed between 0 and 1.
Fourth, we choose the lowest value for $k$ such that for at most $c \cdot 100\%$ of the observations, $k(1 - u_i^2) < t_i$ holds.
Finally, for each observation for which $k(1 - u_i^2) < t_i$, we set $t_i = k(1 - u_i^2)$ and $\delta_i = 0$. For every other observation, we leave $t_i$ and set $\delta_i = 1$.

We evaluate each method with a depth limit of four on five generated data sets for each combination of $n \in \{100, 200, 500, 1000, 2000, 5000\}$ and $c \in \{0.1, 0.5, 0.8\}$, each with a corresponding test set of 50,000 instances. 

\begin{figure*}[t!]
    \centering
    \includegraphics{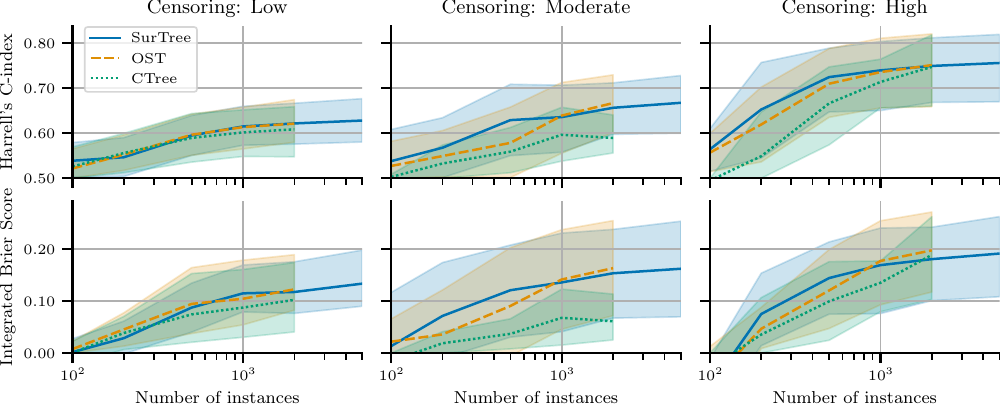}
    \caption{Harrell's C-index and the integrated Brier score on the synthetic data sets (except time-outs).}
    \label{fig:synt_results}
\end{figure*}

\paragraph{Preprocessing}
We use one-hot encoding to encode categorical variables. For categorical variables with more than ten categories, the least frequent categories or combined into an `other' category. Because the dynamic programming approach of SurTree requires binary features, we binarize the numeric features using ten quantiles on all possible thresholds. Identical features and binary features that identify less than 1\% of the data are removed. We evaluate CTree and OST on the numeric data and SurTree on the binarized data.

\subsection{Survival Metrics}
To evaluate the out-of-sample performance of all methods, we compare each method using two common metrics from the literature: \emph{Harrell's C-index} ($H_C$) \citep{harell1982harell_c_index}, and the integrated Brier score ($IB$) \citep{graf1999integrated_brier_score}.

\paragraph{Harrell's C-index}
Harrell's C-index measures the concordance score. Two instances are (dis)concordant if an earlier known death ($t_i < t_j$) for one instance means a (lower) higher risk of death for that instance ($\theta_i > \theta_j$).
When $\theta_i = \theta_j$, the pair is said to have a \textit{tied risk}.
Since for censored observations, the time-to-event is not known, we can only compare pairs for which the instance with an \textit{earlier time} is not censored. 
The number of concordant, discordant, and tied-risk pairs can be calculated using the following formulas respectively:
\begin{align}%
    CC &= \textstyle{\sum_{i,j}} \boldone(t_i < t_j) \boldone(\theta_i > \theta_j) \delta_i \\
    DC &= \textstyle{\sum_{i,j}} \boldone(t_i < t_j) \boldone(\theta_i < \theta_j) \delta_i \\
    TR &= \textstyle{\sum_{i,j}} \boldone(t_i < t_j) \boldone(\theta_i = \theta_j) \delta_i
\end{align}%
Harrell’s C-index is computed as follows:
\begin{equation}
    H_C = \frac{CC + 0.5 \cdot TR}{CC + TR + DC}
\end{equation}

The advantage of Harrell's C-index is that it does not make any parametric assumptions and that it works well for the proportional hazard model as used in this paper. Its disadvantage is that it does not take incomparable pairs into account, which is a problem when censoring is high. Note that a random predictor has an expected score of $H_C = 0.5$.

\paragraph{Integrated Brier score} The Brier score \citep{brier1950brier_score} is commonly used to evaluate probability forecasts, and measures the mean square prediction error. This measure can be used to evaluate survival distributions at a specific point in time. For evaluating the whole distribution, \citet{graf1999integrated_brier_score} developed the integrated Brier score:
\begin{equation}
    IB = \frac{ \sum_i \int_{\ubar{t}}^{t_i} \frac{(1 - \hat{S}_i(t))^2}{\hat{G}(t)}dt + \delta_i \int_{t_i}^{\bar{t}} \frac{(\hat{S}_i(t))^2}{\hat{G}(t_i)} dt }{{|\mathcal{D}|(\bar{t} - \ubar{t})}}
\end{equation}

The integrated Brier score evaluates the Brier score over a time interval, with each time step weighed by the Kaplan-Meier estimator of the censoring distribution $\hat{G}(t)$. 
We compute the integrated Brier score using the test data over the time periods that fall within the 10\% and 90\% quantile of $t_i$ in the test data, given by $\ubar{t}$ and $\bar{t}$ respectively. 
For easier comparison, we report the normalized relative score ($1-IB/IB_0$), with  $IB_0$ the score obtained from the Kaplan-Meier estimator over the whole dataset.

The integrated Brier score also makes no parametric assumptions on the data. Another advantage is that it considers both the censored and non-censored data.

\subsection{Scalability}
\begin{figure}
    \centering
    \includegraphics{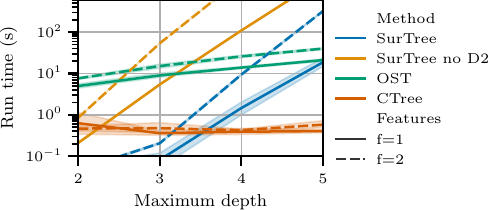}
    \caption{Run time performance for increasing depth, for 3 continuous, 1 binary, and 2 categorical features ($f=1$) or 6 continuous, 2 binary, and 4 categorical features ($f=2$).}
    \label{fig:runtime}
\end{figure}

\begin{table*}[t!]
    \centering
    \begin{tabular}{l cccc ccc ccc}
        \toprule
         &&&&&
         \multicolumn{3}{c}{Harrell's C-index} &
         \multicolumn{3}{c}{Integrated Brier Score} \\
        \cmidrule(lr){6-8}
        \cmidrule(lr){9-11}
        Data set & $|\mathcal{D}|$ & Censoring (\%) & $|\mathcal{F}_{num}|$ & $|\mathcal{F}|$ &
        CTree & OST & SurTree & CTree & OST & SurTree \\ \midrule

Aids2 &
2839 & 38.0\% & 4 & 22 &
\textbf{0.53} & % HC ctree
\textbf{0.53} & % HC ost
\textbf{0.53} & % HC streed
\textbf{0.01} & % IBS ctree
\textbf{0.01} & % IBS ost
0.00 \\ % IBS streed
Dialysis &
6805 & 76.4\% & 4 & 35 &
0.64 & % HC ctree
0.65 & % HC ost
\textbf{0.66} & % HC streed
0.07 & % IBS ctree
\textbf{0.09} & % IBS ost
0.08 \\ % IBS streed
Framingham &
4658 & 68.5\% & 7 & 60 &
0.67 & % HC ctree
0.67 & % HC ost
\textbf{0.68} & % HC streed
0.09 & % IBS ctree
\textbf{0.10} & % IBS ost
\textbf{0.10} \\ % IBS streed
Unempdur &
3241 & 38.7\% & 6 & 45 &
\textbf{0.70} & % HC ctree
0.69 & % HC ost
0.69 & % HC streed
\textbf{0.11} & % IBS ctree
0.10 & % IBS ost
0.10 \\ % IBS streed
Acath &
2258 & 34.0\% & 3 & 21 &
0.59 & % HC ctree
0.58 & % HC ost
\textbf{0.60} & % HC streed
\textbf{0.03} & % IBS ctree
0.02 & % IBS ost
\textbf{0.03} \\ % IBS streed
Csl &
2481 & 89.1\% & 6 & 42 &
\textbf{0.78} & % HC ctree
0.76 & % HC ost
0.75 & % HC streed
\textbf{0.10} & % IBS ctree
\textbf{0.10} & % IBS ost
\textbf{0.10} \\ % IBS streed
Datadivat1 &
5943 & 83.6\% & 5 & 21 &
0.63 & % HC ctree
\textbf{0.64} & % HC ost
0.63 & % HC streed
\textbf{0.08} & % IBS ctree
0.05 & % IBS ost
0.06 \\ % IBS streed
Datadivat3 &
4267 & 94.4\% & 7 & 30 &
0.65 & % HC ctree
0.63 & % HC ost
\textbf{0.66} & % HC streed
0.02 & % IBS ctree
0.02 & % IBS ost
\textbf{0.03} \\ % IBS streed
Divorce &
3371 & 69.4\% & 3 & 5 &
0.52 & % HC ctree
\textbf{0.53} & % HC ost
\textbf{0.53} & % HC streed
0.01 & % IBS ctree
\textbf{0.02} & % IBS ost
\textbf{0.02} \\ % IBS streed
Flchain &
6524 & 69.9\% & 10 & 60 &
\textbf{0.92} & % HC ctree
\textbf{0.92} & % HC ost
\textbf{0.92} & % HC streed
0.65 & % IBS ctree
0.65 & % IBS ost
\textbf{0.66} \\ % IBS streed
Hdfail &
52422 & 94.5\% & 6 & 27 &
- & % HC ctree
-  & % Timeout ost, HC 0.50
\textbf{0.81} & % HC streed
- & % IBS ctree
-  & % Timeout ost, IBS 0.00
\textbf{0.41} \\ % IBS streed
Nwtco &
4028 & 85.8\% & 7 & 17 &
\textbf{0.70} & % HC ctree
\textbf{0.70} & % HC ost
0.69 & % HC streed
0.12 & % IBS ctree
\textbf{0.13} & % IBS ost
\textbf{0.13} \\ % IBS streed
Oldmort &
6495 & 69.7\% & 7 & 33 &
0.64 & % HC ctree
\textbf{0.65} & % HC ost
0.63 & % HC streed
\textbf{0.06} & % IBS ctree
0.05 & % IBS ost
0.05 \\ % IBS streed
Prostatesurvival &
14294 & 94.4\% & 3 & 8 &
\textbf{0.75} & % HC ctree
\textbf{0.75} & % HC ost
\textbf{0.75} & % HC streed
0.09 & % IBS ctree
\textbf{0.10} & % IBS ost
\textbf{0.10} \\ % IBS streed
Rott2 &
2982 & 57.3\% & 11 & 50 &
0.68 & % HC ctree
0.68 & % HC ost
\textbf{0.69} & % HC streed
0.12 & % IBS ctree
\textbf{0.15} & % IBS ost
0.14 \\ % IBS streed
\midrule
Wins per metric & &&&&
6 & % ctree # HC wins
7 & % ost # HC wins
\textbf{10} & % streed # HC wins
6 & % ctree # IBS wins
8 & % ost # IBS wins
\textbf{9} \\ % streed # IBS wins
Average rank & &&&&
2.07 & % ctree # HC rank
2.03 & % ost # HC rank
\textbf{1.83} & % streed # HC rank
2.21 & % ctree # IBS rank
1.97 & % ost # IBS rank
\textbf{1.77} \\ % streed # IBS rank

        \bottomrule
    \end{tabular}
    \caption{Out-of-sample Harrell's C-index and integrated Brier score for data sets from SurvSet \citep{drysdale2022survset} for trees of maximum depth $d=3$. $|\mathcal{F}_{num}|$ is the number of original features. $|\mathcal{F}|$ is the resulting number of binarized features.}
    \label{tab:hc_ibs_survset}
\end{table*}

\paragraph{Synthetic data} To evaluate the scalability of each method, we compare the run time of each algorithm for increasing maximum depth and features. Each method is evaluated twice for five synthetic datasets with $n=5000$ and $c=0.5$. Once for the original setting ($f=1$) with three continuous, one binary, and two categorical features, and once with double the number of features ($f=2$): six continuous, two binary, and four categorical features. After binarization, this results in 39 and 78 binary features, respectively.

Fig.~\ref{fig:runtime} shows that for $f=1$ up to depth 5, and $f=2$ up to depth 4, SurTree has a lower run time than OST. SurTree's run time scales exponentially with increasing maximum depth, whereas OST's run time scales linearly. OST has a relatively high run time for low depth because it randomly restarts its local search several times (by default 100 times) to improve the quality of the solution. In contrast, SurTree immediately finds the globally optimal solution. CTree surprisingly has approximately a constant run time for increasing depth and number of features.

\paragraph{Real data} Despite SurTree being an optimal method, SurTree's average run time for optimizing trees of maximum depth three for the real data sets (including hyper-tuning)
is lower than both CTree and OST. On average, it is more than 100 times faster than OST (geometric mean performance ratio). CTree's worse performance here must be attributed to the cross-validation method.

\paragraph{Depth-two algorithm} Fig.~\ref{fig:runtime} also shows the increase in scalability due to our algorithm for trees of depth two. On average, the depth-two algorithm reduces run time 45 times (geometric mean, not considering time-outs).

\subsection{Out-of-sample results}
\begin{figure}
    \centering
    \includegraphics[width=\columnwidth]{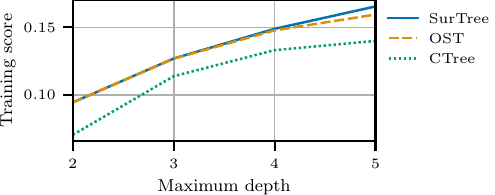}
    \caption{Normalized training loss for CTree, OST, and SurTree, when trained with binarized data.}
    \label{fig:train_score}
\end{figure}

\paragraph{Synthetic data} Fig.~\ref{fig:synt_results} shows the performance on the synthetic data for an increasing number of instances. The results are split for low, moderate, and high censoring.
Since Harrell's C-index only measures performance for instances that are comparable%
%(for non-censored data where the true time-of-event is known, or for instances where the earlier time is non-censored)
, Harrell's C-index is slightly higher for high censoring. %The integrated Brier score is slightly lower for high censoring.
In general, each method performs better with more data, but both OST and CTree time out when $n=5000$.
Furthermore, these results show that both OST and SurTree perform significantly better than CTree, specifically for moderate and high censoring. 
OST and SurTree perform similarly, but a Wilcoxon signed rank test shows that SurTree has a better Harrell's C-index than OST for moderate and high censoring and few instances ($n=200, 500)$.

\paragraph{Real data} 
Table~\ref{tab:hc_ibs_survset} shows the out-of-sample $H_C$ and $IB$ scores for trees with a maximum depth of three. A Wilcoxon signed rank test reveals that both OST and SurTree perform significantly better (95\% confidence) than CTree on both Harrell's C-index and the integrated Brier score. On average, SurTree performs slightly better than OST, but this difference is not statistically significant. Both OST and CTree resulted in one time-out (for Hdfail). For CTree, this is the result of a slow cross-validation algorithm in R based on Harrell's C-index that requires a quadratic number of comparisons. The appendix also shows results for trees with a maximum depth of four.

\paragraph{Training score}
Since SurTree is the first optimal survival tree method, we can now measure in reasonable time how far non-optimal methods are from the optimal solution, when comparing training scores on the same (binarized) data.
Fig.~\ref{fig:train_score} compares the mean training score of CTree, OST, and SurTree on five synthetic training data sets with $n=5000$ and $c=0.5$, without hyper-tuning. 
In this figure, the training score is the normalized loss, with 0 referring to the loss of a single leaf node and 1 referring to zero loss. 
The difference between SurTree and OST is greatest for $d=5$, where SurTree's training score is 4\% better than OST's. The difference between SurTree and CTree is greatest for $d=2$, where SurTree's training score is 34\% higher than CTree's.

\section{Conclusion}
We present SurTree, the first survival tree method with global optimality guarantees. The out-of-sample comparison shows it performs better than an existing greedy heuristic and similar to a state-of-the-art local search approach.
SurTree uses dynamic programming and a special algorithm for trees of depth two resulting in run times even lower than the state-of-the-art local search method that does not provide optimality guarantees.

To improve the prediction quality, future work could explore the effect of fitting a Cox proportional hazards model in each leaf node, instead of only a single constant proportional hazard parameter \citep{cox1972sa_reg}.

\DeclareRobustCommand{\VAN}[3]{#3}
\bibliography{references, manual_references}

\DeclareRobustCommand{\VAN}[3]{#2}

%%%%%%
\newpage

\appendix

\begin{table*}[!t]
    \centering
    \begin{tabular}{l cccc ccc ccc}
        \toprule
         &&&&&
         \multicolumn{3}{c}{Harrell's C-index} &
         \multicolumn{3}{c}{Integrated Brier Score} \\
        \cmidrule(lr){6-8}
        \cmidrule(lr){9-11}
        Data set & $|\mathcal{D}|$ & Censoring (\%) & $|\mathcal{F}_{num}|$ & $|\mathcal{F}|$ &
        CTree & OST & SurTree & CTree & OST & SurTree \\ \midrule

Aids2 &
2839 & 38.0\% & 4 & 22 &
\textbf{0.53} & % HC ctree
\textbf{0.53} & % HC ost
\textbf{0.53} & % HC streed
\textbf{0.01} & % IBS ctree
\textbf{0.01} & % IBS ost
0.00 \\ % IBS streed
Dialysis &
6805 & 76.4\% & 4 & 35 &
0.64 & % HC ctree
\textbf{0.67} & % HC ost
0.65 & % HC streed
0.07 & % IBS ctree
\textbf{0.12} & % IBS ost
0.08 \\ % IBS streed
Framingham &
4658 & 68.5\% & 7 & 60 &
\textbf{0.68} & % HC ctree
0.67 & % HC ost
0.67 & % HC streed
\textbf{0.11} & % IBS ctree
0.09 & % IBS ost
0.10 \\ % IBS streed
Unempdur &
3241 & 38.7\% & 6 & 45 &
0.69 & % HC ctree
0.69 & % HC ost
\textbf{0.70} & % HC streed
\textbf{0.11} & % IBS ctree
\textbf{0.11} & % IBS ost
\textbf{0.11} \\ % IBS streed
Acath &
2258 & 34.0\% & 3 & 21 &
\textbf{0.59} & % HC ctree
\textbf{0.59} & % HC ost
\textbf{0.59} & % HC streed
0.02 & % IBS ctree
0.02 & % IBS ost
\textbf{0.03} \\ % IBS streed
Csl &
2481 & 89.1\% & 6 & 42 &
\textbf{0.78} & % HC ctree
0.76 & % HC ost
0.76 & % HC streed
0.10 & % IBS ctree
0.10 & % IBS ost
\textbf{0.11} \\ % IBS streed
Datadivat1 &
5943 & 83.6\% & 5 & 21 &
0.63 & % HC ctree
\textbf{0.64} & % HC ost
0.63 & % HC streed
\textbf{0.08} & % IBS ctree
0.05 & % IBS ost
\textbf{0.08} \\ % IBS streed
Datadivat3 &
4267 & 94.4\% & 7 & 30 &
\textbf{0.66} & % HC ctree
0.64 & % HC ost
0.65 & % HC streed
0.02 & % IBS ctree
\textbf{0.03} & % IBS ost
\textbf{0.03} \\ % IBS streed
Divorce &
3371 & 69.4\% & 3 & 5 &
0.52 & % HC ctree
\textbf{0.53} & % HC ost
\textbf{0.53} & % HC streed
0.01 & % IBS ctree
\textbf{0.02} & % IBS ost
\textbf{0.02} \\ % IBS streed
Flchain &
6524 & 69.9\% & 10 & 60 &
\textbf{0.92} & % HC ctree
\textbf{0.92} & % HC ost
\textbf{0.92} & % HC streed
\textbf{0.66} & % IBS ctree
\textbf{0.66} & % IBS ost
\textbf{0.66} \\ % IBS streed
Hdfail &
52422 & 94.5\% & 6 & 27 &
-  & % Timeout ctree, HC 0.50
-  & % Timeout ost, HC 0.50
\textbf{0.84} & % HC streed
-  & % Timeout ctree, IBS 0.00
-  & % Timeout ost, IBS 0.00
\textbf{0.44} \\ % IBS streed
Nwtco &
4028 & 85.8\% & 7 & 17 &
0.70 & % HC ctree
\textbf{0.71} & % HC ost
0.70 & % HC streed
\textbf{0.14} & % IBS ctree
\textbf{0.14} & % IBS ost
0.13 \\ % IBS streed
Oldmort &
6495 & 69.7\% & 7 & 33 &
\textbf{0.64} & % HC ctree
-  & % Timeout ost, HC 0.53
\textbf{0.64} & % HC streed
\textbf{0.06} & % IBS ctree
-  & % Timeout ost, IBS 0.01
\textbf{0.06} \\ % IBS streed
Prostatesurvival &
14294 & 94.4\% & 3 & 8 &
\textbf{0.76} & % HC ctree
\textbf{0.76} & % HC ost
\textbf{0.76} & % HC streed
0.10 & % IBS ctree
\textbf{0.11} & % IBS ost
\textbf{0.11} \\ % IBS streed
Rott2 &
2982 & 57.3\% & 11 & 50 &
0.68 & % HC ctree
0.68 & % HC ost
\textbf{0.69} & % HC streed
0.12 & % IBS ctree
\textbf{0.15} & % IBS ost
\textbf{0.15} \\ % IBS streed
\midrule
Wins per metric & &&&&
8 & % ctree # HC wins
8 & % ost # HC wins
\textbf{9} & % streed # HC wins
7 & % ctree # IBS wins
9 & % ost # IBS wins
\textbf{11} \\ % streed # IBS wins
Average rank & &&&&
2.07 & % ctree # HC rank
2.07 & % ost # HC rank
\textbf{1.87} & % streed # HC rank
2.23 & % ctree # IBS rank
2.03 & % ost # IBS rank
\textbf{1.73} \\ % streed # IBS rank

        \bottomrule
    \end{tabular}
    \caption{Out-of-sample Harrell's C-index and integrated brier score for data sets from SurvSet \citep{drysdale2022survset} for trees of maximum depth $d=4$. $|\mathcal{F}_{num}|$ is the number of original features. $|\mathcal{F}|$ is the resulting number of binarized features.}
    \label{tab:hc_ibs_survset_d4}
\end{table*}

\section{Loss function}
Let $h\in T$ be the leaf nodes of a tree $T$, $\hat{\theta}_h$ the $\theta$ estimate for leaf node $h$ and $\mathcal{D}_h$ the instances that end up in this leaf node. Then, as in \citep{leblanc1992sa_rel_risk}, the likelihood function of a survival tree can be formulated as:

\begin{equation}
    \label{app_eq:likelihood}
    L = \prod_{h\in T} \prod_{(t_i, \delta_i, \textbf{fv}_i) \in \mathcal{D}_h} (\hat{\theta}_h \hat{\lambda}(t_i))^{\delta_i} e^{-\hat{\theta}_h\hat{\Lambda}(t_i)}
\end{equation}

Here $\hat{\Lambda}(t)$ is an estimate of the baseline cumulative hazard function. As similarly done in \citep{bertsimas2022survival}, we estimate it using the Nelson-Aalen estimator, which is a first iteration estimator according to \citet{leblanc1992sa_rel_risk}.

Eq.~\eqref{app_eq:likelihood} results in the partial log-likelihood function $LL$ for a single leaf node:

\begin{equation}
\label{app_eq:log_likelihood}
    LL(\mathcal{D}, \hat{\theta}) = \sum_{(t_i, \delta_i, \textbf{fv}_i) \in \mathcal{D}} 
    \delta_i \log \hat{\lambda}(t_i)
    + \delta_i \log \hat{\theta} - \hat{\Lambda}(t_i) \hat{\theta}
\end{equation}

Taking the derivative with respect to $\hat{\theta}$ results in the following:

\begin{equation}
    \frac{dLL(\mathcal{D}, \hat{\theta})}{d\hat{\theta}} = 
    \sum_{(t_i, \delta_i, \textbf{fv}_i) \in \mathcal{D}} \left( \frac{\delta_i}{\hat{\theta}} - \hat{\Lambda}(t_i) \right)
\end{equation}

Setting the derivative to zero yields the maximum likelihood value for $\hat{\theta}$ in this leaf node:

\begin{equation}
    \label{app_eq:surtree_theta_function}
    \hat{\theta} = \frac{\sum_{(t_i, \delta_i, \textbf{fv}_i) \in \mathcal{D}} \delta_i}{\sum_{(t_i, \delta_i, \textbf{fv}_i) \in \mathcal{D}} \hat{\Lambda}(t_i)}    
\end{equation}

When the data set consists of only one instance $i$, Eq.~\eqref{app_eq:surtree_theta_function} gives the \emph{saturated} coefficient $\hat{\theta}^{sat}_i$ that perfectly maximizes the likelihood for this instance alone:
\begin{equation}
\hat{\theta}^{sat}_i = \frac{\delta_i}{\hat{\Lambda}(t_i)}
\end{equation}%

The log-likelihood for one instance is given by Eq.~\eqref{app_eq:log_likelihood} when the data set consists of one instance. By this equation, the loss for a single instance $i$ is then defined as the difference between the log-likelihood of the instance's $\hat{\theta}^{sat}_i$ and the log-likelihood of the instance's leaf node $\hat{\theta}$. 

\begingroup
\allowdisplaybreaks
\begin{equation}
\begin{aligned}
    \mathcal{L}(\hat{\theta}) =& \left(\delta_i \log \hat{\lambda}(t_i)
    + \delta_i \log \frac{\delta_i}{\hat{\Lambda}(t_i)} - \hat{\Lambda}(t_i)\frac{\delta_i}{\hat{\Lambda}(t_i)} \right) \\
    &- \left( \delta_i \log \hat{\lambda}(t_i)
    + \delta_i \log \hat{\theta} - \hat{\Lambda}(t_i) \hat{\theta} \right) \\
    =& \hat{\Lambda}(t_i) \hat{\theta} + \delta_i \log \frac{\delta_i}{\hat{\Lambda}(t_i)} - \delta_i - \delta_i \log \hat{\theta} \\
    %=& \hat{\Lambda}(t_i) \hat{\theta} + \delta_i \log \delta_i - \delta_i \log \hat{\Lambda}(t_i) - \delta_i - \delta_i \log \hat{\theta} \\
    =& \hat{\Lambda}(t_i) \hat{\theta} - \delta_i \log \hat{\Lambda}(t_i) - \delta_i \log \hat{\theta} - \delta_i
\end{aligned}
\end{equation}
\endgroup
The last step in this derivation uses the fact that $\delta_i$ is binary and therefore $\delta_i \log \delta_i$ is always zero.
Summing the loss of the instances in a leaf node yields the loss function in the main text:
\begin{multline}
\label{app_eq:surtree_error_function}
\mathcal{L}(\mathcal{D}, \hat{\theta}) = \\
\sum_{(t_i, \delta_i, \textbf{fv}_i) \in \mathcal{D}} \Big( \hat\Lambda(t_i) \hat{\theta} -\delta_i\log\hat\Lambda(t_i) - \delta_i\log\hat{\theta} - \delta_i \Big)
\end{multline}

To prevent infinite loss when $\hat{\theta} = 0$, \citet{leblanc1992sa_rel_risk} set $\hat{\theta} = 1 / (2 \sum_{i} \hat{\Lambda}(t_i))$ when no events are recorded in a leaf node.

The loss for the whole tree can be computed by summing the loss of each leaf node.

\section{Depth-two solver}
Let $X$ denote the cost tuple $(\operatorname{ES}, \operatorname{HS}, \operatorname{NLHS})$ for the depth-two algorithm, consisting of the \emph{event sum} $\operatorname{ES}$, the \emph{hazard sum} $\operatorname{HS}$ and the \emph{negative log hazard sum} $\operatorname{NLHS}$. Let $C(\operatorname{ES}, \operatorname{HS}, \operatorname{NLHS})$ compute the loss from the tuple $X$:
\begin{equation}
    C(\operatorname{ES}, \operatorname{HS}, \operatorname{NLHS}) = \operatorname{NLHS} - \operatorname{ES} \log\bigg(\frac{\operatorname{ES}}{\operatorname{HS}}\bigg)
\end{equation}

Then the depth-two algorithm is given by the pseudo-code in Algorithm~\ref{alg:depth_two}. In this algorithm, the $X$ tuples are summed using element-wise addition. The first step is the pre-computation of the $X$ tuples. The second step is going over all possible branching nodes and computing the loss from the pre-computed values. The final step is to return the loss of the root node that has the minimal sum of left and right loss.

The values for $X$ that are not pre-computed can be derived using the following formulas:

\begin{align}
    X(\overline{f_i}) &= X - X(f_i) \\
    X(f_i, \overline{f_j}) &= X(f_i) - X(f_i, f_j) \\
    X(\overline{f_i}, f_j) &= X(f_j) - X(f_i, f_j) \\
    X(\overline{f_i}, \overline{f_j}) &= X - X(f_i) - X(f_j) + X(f_i, f_j)
\end{align}

\setlength{\algomargin}{0pt}
\begin{algorithm}[tb!]
\caption{First, pre-compute the tuples $X$ describing the event, hazard, and negative log hazard sum. Then find for every feature the best loss for the left and right subtree $BestL.\mathcal{L}$ and $BestR.\mathcal{L}$. Finally, return the loss of the best tree of at most depth two.}
\label{alg:depth_two}
\DontPrintSemicolon
\SetInd{.5em}{.5em}
$X \leftarrow (0,0,0)$\;
$X(f_i) \leftarrow (0,0,0) \quad \forall f_i \in \mathcal{F}$\;
$X(f_i, f_j) \leftarrow (0,0,0) \quad \forall f_i, f_j, \in \mathcal{F}$ s.t. $i < j$ \;
\For{$(t, \delta, \mathbf{fv}) \in \mathcal{D}$} {
    $X \leftarrow X + (\delta, \hat{\Lambda}(t), -\delta \log \hat{\Lambda}(t))$\;
    \For{$f_i \in \mathbf{fv}$} {
        $X(f_i) \leftarrow X(f_i) + (\delta, \hat{\Lambda}(t), -\delta \log \hat{\Lambda}(t))$\;
        \For{$f_j \in \mathbf{fv}$, s.t. $i < j$} {
            $X(f_i, f_j) \leftarrow X(f_i, f_j) + (\delta, \hat{\Lambda}(t), -\delta \log \hat{\Lambda}(t))$\;
        }
    }
}
\For{$f_i \in \mathcal{F}$} {
    \For{$f_j \in \mathcal{F}$} {
        $\mathcal{L}_L = C(X(\bar{f_i}, f_j)) + C(X(\bar{f_i}, \bar{f_j}))$\;
        $\mathcal{L}_R = C(X(f_i, f_j)) + C(X(f_i, \bar{f_j}))$\;
        \If{$BestL.\mathcal{L}(f_i) > \mathcal{L}_L$} {
            $BestL.\mathcal{L}(f_i) \leftarrow \mathcal{L}_L$\;
        }
        \If{$BestR.\mathcal{L}(f_i) > \mathcal{L}_R$} {
            $BestR.\mathcal{L}(f_i) \leftarrow \mathcal{L}_R$\;
        }
    }
}
\Return $\min_{f_i \in \mathcal{F}} BestL.\mathcal{L}(f_i) + BestR.\mathcal{L}(f_i)$\;

\end{algorithm}

\section{Distributions in Ground Truth Trees}
\label{apdx:tree_distributions}

To determine the time-to-event for each instance during the synthetic dataset generation, a ground truth tree is generated for each dataset. As also done in \citep{bertsimas2022survival}, each leaf node is assigned a random distribution from the list below. Each option has an equal probability of being assigned.

\begin{itemize}
    \item $\text{Exponential}(\lambda)$, with $\lambda \in \{0.3,$ $0.4$, $0.6$, $0.8$, $0.9$, $1.15$, $1.5$, $1.8\}$
    \item $\text{Weibull}(k, \lambda)$, with $(k, \lambda) \in \{(0.8, 0.4)$, $(0.9, 0.5)$, $(0.9, 0.7)$, $(0.9, 1.1)$, $(0.9, 1.5)$, $(1.0, 1.1)$, $(1.0, 1.9)$, $(1.3, 0.5)\}$
    \item $\text{Lognormal}(\mu, \sigma^2)$, with $(\mu, \sigma^2) \in \{(0.1, 1)$, $(0.2, 0.75)$, $(0.3, 0.3)$, $(0.3, 0.5)$, $(0.3, 0.8)$, $(0.4, 0.32)$, $(0.5, 0.3)$, $(0.5, 0.7)\}$
    \item $\text{Gamma}(k, \theta)$, with $(k, \theta) \in \{(0.2, 0.75)$, $(0.3, 1.3)$, $(0.3, 2.0)$, $(0.5, 1.5)$, $(0.8, 1.0)$, $(0.9, 1.3)$, $(1.3, 0.9)$, $(1.5, 0.7)\}$
\end{itemize}

%\newpage
\section{Extended Experiment Results}

Table~\ref{tab:hc_ibs_survset_d4} shows the out-of-sample performance of each method for trees up to depth four. A Wilcoxon signed rank test points out that no significant differences between the methods can be observed, except for one: SurTree has a significantly better integrated Brier score than CTree ($p < 5\%)$.

%%%%%%

\end{document}